\documentclass[11pt]{article}

\usepackage[final]{acl}

\usepackage{times}
\usepackage{latexsym}

\usepackage[T1]{fontenc}

\usepackage[utf8]{inputenc}

\usepackage{microtype}

\usepackage{inconsolata}

\usepackage{graphicx}

\usepackage{xcolor}
\usepackage{amsmath}
\usepackage{tabularx}
\usepackage{booktabs}
\usepackage{colortbl}
\usepackage{wrapfig}
\usepackage{color}
\usepackage{verbatim}
\usepackage{ulem}
\usepackage{graphicx}
\usepackage{amsfonts}

\title{CtrlDiff: Boosting Large Diffusion Language Models with Dynamic Block Prediction and Controllable Generation}

\author{Chihan Huang$^{1,}$\thanks{Work Done during the visit at Peking University.} \\
  $^1$Nanjing University of Science and Technology \\
  \texttt{huangchihan@njust.edu.cn} \\\And
  Hao Tang$^{2,}$\thanks{Corresponding author.} \\
  $^2$Peking University \\
  \texttt{haotang@pku.edu.cn}}

\begin{document}
\maketitle
\begin{abstract}
Although autoregressive models have dominated language modeling in recent years, there has been a growing interest in exploring alternative paradigms to the conventional next-token prediction framework. Diffusion-based language models have emerged as a compelling alternative due to their powerful parallel generation capabilities and inherent editability. However, these models are often constrained by fixed-length generation. A promising direction is to combine the strengths of both paradigms, segmenting sequences into blocks, modeling autoregressive dependencies across blocks while leveraging discrete diffusion to estimate the conditional distribution within each block given the preceding context. Nevertheless, their practical application is often hindered by two key limitations: rigid fixed-length outputs and a lack of flexible control mechanisms. In this work, we address the critical limitations of fixed granularity and weak controllability in current large diffusion language models. We propose CtrlDiff, a dynamic and controllable semi-autoregressive framework that adaptively determines the size of each generation block based on local semantics using reinforcement learning. Furthermore, we introduce a classifier-guided control mechanism tailored to discrete diffusion, which significantly reduces computational overhead while facilitating efficient post-hoc conditioning without retraining. Extensive experiments demonstrate that CtrlDiff sets a new standard among hybrid diffusion models, narrows the performance gap to state-of-the-art autoregressive approaches, and enables effective conditional text generation across diverse tasks.
\end{abstract}

\section{Introduction}
\label{section1}

The concept of forecasting future events has long been central to artificial intelligence. Among various modeling paradigms, auto-regressive (AR) models have achieved remarkable success through the "predict-the-next-token" strategy \cite{vaswani2017-attention, brown2020-language, chowdhery2023-palm}, attaining state-of-the-art results in domains such as code generation \cite{roziere2024-code, ahmad2025-opencode}, complex reasoning \cite{trinh2024-solving, wei2022-chain}, and medical applications \cite{liu2025-application, busch2025-current}. However, AR models are inherently constrained by their left-to-right sampling paradigm, which enforces a strict sequential dependency during both training and inference \cite{hoogeboom2022-autoregressive, shih2022-training}. This structure limits flexibility and often leads to cumulative error propagation, where early generation mistakes compound throughout the sequence \cite{bengio2015-scheduled}. In contrast, diffusion models \cite{sohl2015-deep, ho2020-denoising}, initially designed for continuous data like images and videos, have achieved remarkable success with advances such as Stable Diffusion \cite{rombach2022-high, podell2024-sdxl, esser2024-scaling} and Sora. More recently, discrete diffusion models \cite{austin2021-structured} have extended this paradigm to inherently discrete domains, including text \cite{li2022-diffusionlm}, molecular structures \cite{gruver2023-protein}, and genomic sequences \cite{li2024-discdiff}, enabling sequence generation via iterative denoising from fully masked inputs.

Unlike AR models, discrete diffusion models generate sequences in parallel by progressively denoising a fully masked input, offering bidirectional controllability and faster sampling \cite{xu2025-energy}. However, existing discrete diffusion models remain limited to fixed-length generation and still underperform AR models in quality. Moreover, gradient-based conditioning methods effective in continuous domains cannot be directly applied to discrete data due to their non-differentiable nature.

Although hybrid models \cite{arriola2025-block} that integrate autoregressive decoding with block-wise diffusion offer a promising middle ground, they face two major limitations: (1) each block size requires retraining a separate model, and the block size must remain fixed during sampling, limiting fine-grained control and efficiency; (2) they do not support conditional generation, restricting their use in guided or controllable generation tasks.

We propose CtrlDiff, a dynamic and controllable semi-autoregressive framework that bridges expressive generation and structural flexibility. Leveraging intra-block independence, our classifier guidance enables controllable generation without retraining or fine-tuning. Furthermore, a reinforcement learning–based dynamic block size predictor adjusts block sizes during generation, balancing semantic precision with computational efficiency.

Our contributions can be summarized as follows:

\begin{enumerate}

\item We introduce CtrlDiff, a hybrid architecture that interpolates between the autoregressive and diffusion models. An efficient classifier guidance strategy is presented, which significantly reduces computational overhead while enabling post-hoc control over generated text conditioned on target attributes.

\item We propose a reinforcement learning-based dynamic block prediction mechanism that allows CtrlDiff to adaptively select block sizes according to sequence semantics to improve fine-grained generation and efficiency.

\item Extensive experiments demonstrate that CtrlDiff outperforms previous diffusion and semi-autoregressive models in both likelihood modeling and conditional generation, offering improved fluency, control accuracy, and semantic consistency.

\end{enumerate}

\section{Related Works}
\label{section2}

\subsection{Diffusion Models}\label{section2.1}

Diffusion models have achieved remarkable success in high-quality image generation \cite{song2019-generative, dhariwal2021-diffusion}. Built on stochastic processes and Markov chains, they synthesize data through a forward noising and a learned reverse denoising process \cite{zheng2025-masked}. While originally designed for continuous data, recent works have extended them to discrete domains.

\cite{austin2021-structured} proposed D3PM, introducing a structured categorical corruption process and deriving a discrete ELBO. \cite{lou2024-discrete} developed SEDD, which employs a score-entropy loss to estimate marginal ratio distributions, extending score matching to discrete spaces. Building on D3PM, \cite{sahoo2024-simple} introduced a \texttt{[MASK]} token as an absorbing state, modeling masked–unmasked transitions via a masked language model, thereby improving both theoretical rigor and empirical performance.

\subsection{Autoregressive Modeling}\label{section2.2}

Auto-regressive models have dominated the landscape of language modeling in recent years, producing numerous remarkable achievements \cite{brown2020-language}. These models are predicated on the principle that the generation of each token is conditioned on all previously generated tokens. Formally, given a sequence $\mathbf{x} = (\mathbf{x}^1, \ldots, \mathbf{x}^L)$ sampled from a data distribution $q(\mathbf{x})$, an AR model assumes that the generation of the $\ell$-th token $\mathbf{x}^\ell$ depends exclusively on its prefix $(\mathbf{x}^1, \ldots, \mathbf{x}^{\ell-1})$. Once a token $\mathbf{x}^\ell$ is generated, it is appended to the prefix to form $(\mathbf{x}^1, \ldots, \mathbf{x}^{\ell})$, which is then used to sample the next token $\mathbf{x}^{\ell+1}$. Consequently, the joint likelihood of the sequence can be factorized as:
\begin{equation}
p(\mathbf{x}^1, \ldots, \mathbf{x}^L) = \prod_{\ell=1}^L p(\mathbf{x}^{\ell+1} | \mathbf{x}^1,  \ldots, \mathbf{x}^{\ell}).
\end{equation}

Each conditional probability term $p(\cdot | \cdot)$ on the right-hand side is modeled by a neural network. By incorporating causal masking within the attention mechanism of Transformers, AR models are capable of both training and sampling in a next-token prediction paradigm.

\subsection{Guidance Techniques}\label{section2.3}

In continuous diffusion models, \cite{dhariwal2021-diffusion} introduced classifier and classifier-free guidance, establishing the basis for effective conditional generation. In diffusion language models, Diffusion-LM \cite{li2022-diffusionlm} maps discrete tokens to continuous embeddings, enabling classifier guidance in a latent space, while SSD-LM \cite{han2023-ssdlm} operates directly in vocabulary space, leveraging pretrained classifiers without embeddings. \cite{nisonoff2025-unlocking} further exploited continuous-time Markov chains, perturbing only one dimension per step for fine-grained control.

In autoregressive models, PPLM \cite{dathathri2020-plug} introduced gradient-based attribute steering, and FUDGE \cite{yang2021-fudge} used an attribute predictor to adjust token probabilities during generation. More recently, \cite{sanchez2024-stay} proposed interpolating conditional and unconditional logits at inference, achieving more faithful controllable generation.

\section{Methodology}\label{section3}

\subsection{Diffusion Language Model}\label{section3.1}

\subsubsection{Discrete Diffusion Model}\label{section3.1.1}

Although diffusion models were originally developed for continuous data, recent advances have extended their applicability to discrete data domains, such as text, molecules, and genetic sequences \cite{aaron2024-discrete, shi2024-simplified}. This breakthrough has enabled non-autoregressive generation, where, unlike autoregressive models that predict tokens sequentially, discrete diffusion methods generate entire sequences in parallel. This is achieved by first corrupting clean sequences through a stochastic noising process, and then iteratively denoising them using a learnable reverse process.

Specifically, let us represent a discrete random variable with $K$ categories as a one-hot vector that belongs to the set $\mathcal{V} = \{\mathbf{x} \in \{0,1\}^K: \sum_{i=1}^K \mathbf{x}_i = 1 \} \subset \Delta^K$, where $\Delta^K$ denotes the $K$-simplex. Let $\mathbf{1} = \{1\}^K$ denote the all-ones vector, and let $\odot$ denote the Hadamard product. We designate the $K$-th category as a special token \texttt{[MASK]}, and denote its corresponding one-hot vector as $\mathbf{m} \in \mathcal{V}$.  

Assume a token sequence $\mathbf{x}_0 = \left(\mathbf{x}_0^1, \ldots, \mathbf{x}_0^L\right)$ of length $L$, where each token satisfies the aforementioned conditions. D3PM \cite{austin2021-structured} introduces a structured noising process under the Markov assumption by utilizing a transition matrix $\left[\mathbf{Q}_{t \mid s}\right]_{ij} = q(\mathbf{x}_t = j \mid \mathbf{x}_s = i)$, which governs the corruption dynamics of the sequence:
\begin{equation}
q(\mathbf{x}_t \mid \mathbf{x}_s) = \mathrm{Cat} \left(\mathbf{x}_t; \mathbf{p} = \mathbf{Q}_{t \mid s} \mathbf{x}_s \right),
\end{equation}
where $\mathrm{Cat}(\mathbf{x}; \mathbf{p})$ denotes a categorical distribution over the one-hot vector $\mathbf{x}$, parameterized by the probability vector $\mathbf{p} \in \Delta^K$. Building upon the D3PM framework, MDLM \cite{sahoo2024-simple} focuses on a specific class of noising processes that utilize an absorbing state, leading to a more efficient and streamlined algorithm. Their noising process is defined as follows:
\begin{equation}
q\left(\mathbf{x}_t \mid \mathbf{x}_0\right) = \mathrm{Cat} \left(\mathbf{x}_t; \alpha_t\mathbf{x}_0+\left(1-\alpha_t\right)\boldsymbol{\pi}\right),
\end{equation}
where $\alpha \in \left[0, 1\right]$ denotes a noise schedule that increases monotonically with time step $t$. Let $\alpha_{t \mid s} = \alpha_t / \alpha_s$ denote the relative noise ratio between time steps $t$ and $s$. Under this formulation, the reverse process is defined as:
\begin{equation}
\resizebox{0.87\hsize}{!}{$
\begin{array}{l}
q\left(\mathbf{x}_{s} \mid \mathbf{x}_{t}, \mathbf{x}_{0}\right)=\operatorname{Cat}\left(\mathbf{x}_{s} ;\right. \\
\quad \left.\frac{\left[\alpha_{t \mid s} \mathbf{x}_{t}+\left(1-\alpha_{t \mid s}\right) \mathbf{1} \boldsymbol{\pi}^{\top} \mathbf{x}_{t}\right] \odot\left[\alpha_{s} \mathbf{x}_{0}+\left(1-\alpha_{s}\right) \boldsymbol{\pi}\right]}{\alpha_{t} \mathbf{x}_{t}^{\top} \mathbf{x}_{0}+\left(1-\alpha_{t}\right) \mathbf{x}_{t}^{\top} \boldsymbol{\pi}}\right) .
\end{array}
$}
\end{equation}

MDLM employs a one-hot vector corresponding to the special \texttt{[MASK]} token, with $\boldsymbol{\pi} = \texttt{[MASK]}$. During the noising process, once a token is replaced by the \texttt{[MASK]} token, it remains fixed as such throughout the remainder of the corruption steps. Conversely, in the reverse process, once a \texttt{[MASK]} token is unmasked (i.e., replaced by a valid token), its identity remains unchanged for the rest of the generation trajectory.

\subsubsection{Diffusion Autoregressive Hybrid Model}\label{section3.1.2}

Diffusion autoregressive hybrid model has been proposed to combine the strengths of both paradigms by partitioning a text sequence into several blocks. It defines an auto-regressive distribution across blocks while employing a diffusion model within each block \cite{arriola2025-block}. Specifically, the token sequence $\mathbf{x}$ is divided into $B$ blocks of fixed length $L^{\prime}$, such that $B = L / L^{\prime}$ is an integer. For brevity, we denote each block $\mathbf{x}^{(b-1)L^{\prime}:bL^{\prime}}$ as $\mathbf{x}^b$, representing the token subsequence that spans positions $(b-1)L^{\prime}$ to $bL^{\prime}$. The overall likelihood then factorizes across blocks as:
\begin{equation}
\log p_\theta \left(\mathbf{x}\right) = \sum_{b=1}^B \log p_{\theta} \left( \mathbf{x}^b \mid \mathbf{x}^{<b} \right).
\end{equation}
Each term $p_\theta(\mathbf{x}^b \mid \mathbf{x}^{<b})$ is modeled using a discrete diffusion process within a block of fixed length $L^{\prime}$. Although directly computing $p_\theta(\mathbf{x}_{t-1} \mid \mathbf{x}_t)$ is feasible, to reduce computational overhead, the approach proposed by \cite{ho2020-denoising} and further extended by \cite{hoogeboom2021-argmax} is adopted. This approach utilizes a neural network to predict $p_\theta(\mathbf{x}^b \mid \mathbf{x}_t^b)$, which is then combined with the true posterior $q(\mathbf{x}_s^b \mid \mathbf{x}_t^b, \mathbf{x}^b)$. Specifically, the reverse process is parameterized and constrained within block $b$ as follows:
\begin{equation}
\begin{aligned}
& \quad p_\theta \left(\mathbf{x}_s^b \mid \mathbf{x}_t^b, \mathbf{x}^{<b}\right) \\
& = \sum_{\mathbf{x}^b} q(\mathbf{x}_s^b \mid \mathbf{x}_t^b, \mathbf{x}^b) p_\theta \left(\mathbf{x}^b \mid \mathbf{x}_t^b, \mathbf{x}^{<b}\right).
\end{aligned}
\end{equation}

\subsection{Classifier Guidance}\label{section3.2}

Applying guidance to discrete diffusion models presents significant challenges, primarily due to the non-differentiability of the discrete variables $\mathbf{x}_t$ involved in the guidance mechanism \cite{schiff2025-simple}. In this context, we formalize the guidance as a conditional probability distribution $p(y \mid \mathbf{x}) \in \Delta^N$, where $y \in \{1, \ldots, N\}$ denotes one of the $N$ possible class labels, and $\gamma$ is a temperature parameter that modulates the strength of the guidance.

For a token sequence $\mathbf{x}_t$ with length $L$ segmented into $B$ blocks as defined in Section \ref{section3.1.2}, the reverse process distribution at time step $t$ for block $b$, under classifier guidance, is modified as follows:
\begin{equation}
\resizebox{0.87\hsize}{!}{$
\begin{aligned}
& \quad p_{\gamma} \left(\mathbf{x}_s^b \mid \mathbf{x}_t^b, y, \mathbf{x}^{<b} \right) \\
& \propto p_{\theta} \left(\mathbf{x}_s^b \mid \mathbf{x}_t^b, \mathbf{x}^{<b} \right) \cdot p_{\xi} \left(y \mid \mathbf{x}_s^b, \mathbf{x}_t^b, \mathbf{x}^{<b}\right)^{\gamma},
\end{aligned}
$}
\end{equation}
where $p_{\theta} \left(\mathbf{x}_s^b \mid \mathbf{x}_t^b, \mathbf{x}^{<b} \right)$ is defined by the standard diffusion reverse process. Therefore, the primary objective is to construct a predictive network $p_{\xi} \left(y \mid \mathbf{x}_s^b, \mathbf{x}_t^b, \mathbf{x}^{<b}\right)$, which estimates the class-conditional distribution required to incorporate guidance into the discrete diffusion framework.

Nonetheless, directly computing the term $p_{\xi} \left(y \mid \mathbf{x}_s^b, \mathbf{x}_t^b, \mathbf{x}^{<b}\right)$ during the reverse sampling process of discrete diffusion is infeasible, as it requires evaluating all $K^L$ possible configurations of $\mathbf{x}_s^b$, resulting in prohibitively high computational costs. To address this, we introduce the following strategy.

\paragraph{Intra-block Independence} Given that the main objective of the large diffusion language model is to improve efficiency by generating all positions within a block in parallel, it is reasonable to assume conditional independence among positions within a block. Specifically, we approximate the classifier as $p_{\xi}\left(y \mid \mathbf{x}_s^b, \mathbf{x}^{<b}\right) = \prod_{\ell=1}^{L} p_{\xi} \left(y \mid \mathbf{x}_s^{b, \ell}, \mathbf{x}^{<b}\right)$, where $\mathbf{x}_s^{b, \ell}$ denotes the token at position $\ell$ in block $b$. Under this assumption, the normalization term can be decomposed accordingly:
\begin{equation}
\resizebox{0.87\hsize}{!}{$
\begin{aligned}
Z & = \sum_{\mathbf{x}_s^b} p_\theta \left(\mathbf{x}_s^b | \mathbf{x}_t^b, \mathbf{x}^{<b}\right) \cdot p_\xi \left(y | \mathbf{x}_s^b, \mathbf{x}_t^b, \mathbf{x}^{<b} \right)^{\gamma} \\
& = \prod_{\ell=1}^{L} \sum_{\mathbf{x}_s^{b,\ell}} p_\theta \left(\mathbf{x}_s^{b,\ell} | \mathbf{x}_t^b, \mathbf{x}^{<b}\right) \cdot p_\xi \left(y | \mathbf{x}_s^{b,\ell}, \mathbf{x}_t^{b}, \mathbf{x}^{<b}\right)^{\gamma}.
\end{aligned}
$}
\end{equation}

For each term $p_{\xi}\left(y \mid \mathbf{x}_{s}^{b,\ell}, \mathbf{x}_{t}^{b}, \mathbf{x}^{<b}\right)$ representing classifier probability, directly computing the joint distribution remains computationally intractable due to the exponential complexity. However, we observe that the classifier’s prediction for label $y$ is primarily influenced by the current modified position $\ell$. Therefore, we approximate the classifier probability by $p_{\xi}(y \mid \mathbf{\hat{x}}_{t \mid s_\ell}^{b}, \mathbf{x}^{<b})$, where $\mathbf{\hat{x}}_{t \mid s_\ell}^{b}$ denotes the sequence obtained by replacing the $\ell$-th token in $\mathbf{x}_{t}^{b}$ with $\mathbf{x}_{s}^{b,\ell}$. This simplification allows us to significantly reduce the computational burden. After training the classifier $p_{\xi}$ on noised token vectors, the sampling equation for classifier-guided generation can be reformulated as follows:
\begin{equation}
\resizebox{0.87\hsize}{!}{$
\begin{aligned}
& \quad p_\theta^\gamma \left(\mathbf{x}_s^b | \mathbf{x}_t^b, \mathbf{x}^{<b}, y\right) \\
& = \prod_{\ell=1}^{L} \frac{p_\theta \left(\mathbf{x}_s^{b,\ell} | \mathbf{x}_t^b, \mathbf{x}^{<b}\right) \cdot p_\xi \left(y | \mathbf{\hat{x}}_{t|s_\ell}^{b}, \mathbf{x}^{<b}\right)^\gamma}{\sum_{\mathbf{\hat{x}}_{t|s_\ell}^{b}} p_\theta \left(\mathbf{x}_s^{b,\ell} | \mathbf{x}_t^b, \mathbf{x}^{<b}\right) \cdot p_\xi \left(y | \mathbf{\hat{x}}_{t|s_\ell}^{b,\ell}, \mathbf{x}^{<b}\right)^\gamma}.
\end{aligned}
$}
\end{equation}

\subsection{Dynamic Length Prediction}\label{section3.3}

Given the varying semantic complexity across different segments of text, it is intuitively more reasonable to adopt a dynamic block length. When the block is excessively long, it may introduce an excessive amount of contextual information, potentially leading to the loss of local details. Conversely, if the block is too short, it can increase computational costs while compromising the fluency of the generated text. By dynamically selecting the block length, the model can adaptively choose a smaller block length for semantically complex regions, thereby generating more fine-grained responses. Meanwhile, for simpler segments, a larger block length can be employed to enhance sampling efficiency without sacrificing output quality.  

\paragraph{State Extraction} Since the objective of determining block length is to strike a balance between generation efficiency and text quality, we formalize this as a decision-making problem. After generating a given block, the model dynamically predicts the length of the subsequent block based on its contextual information.

Specifically, it is necessary to collect a set of data comprising states $\mathcal{S}$, actions $\mathcal{A}$, and rewards $\mathcal{R}$ to facilitate the training of the policy network $\pi_{\phi}$. Here, $\mathcal{S}$ denotes the state space of the environment and serves as the input to the policy network $\pi_{\phi}$. Let $\mathbf{s}_b \in \mathcal{S}$ represent the state at block $b$, defined as follows:
\begin{equation}
\resizebox{0.87\hsize}{!}{$
\mathbf{s}_b=\left[\mathrm{WeightPool} \left(\mathbf{h}^{b-M:b-1}\right), \mathrm{Entropy}\left(\mathbf{x}^{b-M:b-1}\right)\right],
$}
\end{equation}
where $\mathbf{H} = \mathbf{h}^{b-M:b-1}$ denotes the hidden states of the preceding $M$ blocks before the $b$-th block, which can be extracted using a pre-trained large language model. The operator $\mathrm{Entropy}\left(\cdot \right)$ computes the information entropy, serving as a measure of the information density within a block. The function $\mathrm{WeightPool}\left(\cdot \right)$ represents a weighted pooling layer, formally defined as:
\begin{equation}
\mathrm{WeightPool}(\mathbf{H})=\sum_{i=1}^M\alpha_i\mathbf{h}^{b-M-1+i}.
\end{equation}
The coefficient $\alpha_i$ can be manually specified or learned adaptively. In the latter case, it is obtained by applying a learnable weight to $\mathbf{h}^i$, followed by a normalization process.

\paragraph{Policy Network} Set $\mathcal{A}$ denotes the action space, which corresponds to the set of possible block lengths that can be selected. For each block $b$, we first aggregate the preceding local patterns by stacking representations along the temporal dimension to get $\mathbf{f}^b=\mathrm{MaxPool}\left(\mathrm{Conv1D}\left(\mathbf{H}\right)\right)$. Subsequently, we concatenate this representation with the corresponding local entropy to get the final state $\mathbf{s} = \mathrm{MLP}\left(\left[\mathbf{f}^b; \mathrm{Entropy}\left(\mathbf{x}^{b-M:b-1}\right)\right]\right)$.Then we feed the final state into the policy network $\pi_{\phi}$ to obtain the probability distribution $\mathbf{p}$ over actions:
\begin{equation}
\resizebox{0.87\hsize}{!}{$
\mathbf{p}=\pi_\phi\left(\mathbf{a}^b \mid \mathbf{s}_b\right) = \mathrm{Softmax}\left(\mathbf{W}\cdot \mathbf{s} + \mathbf{b}\right).
$}
\end{equation}
Based on the probability distribution $\mathbf{p}$, the policy network $\pi_{\phi}$ selects an action $\mathbf{a}^b \in \mathcal{A}$, which determines the length of the next block as $L_b$. In order to jointly optimize for both high generation quality and computational efficiency, we define the reward function as follows:
\begin{equation}
\resizebox{0.87\hsize}{!}{$
\begin{aligned}
& R\left(s_{b}, a\right)=\underbrace{\lambda_{1} \max\left\{\frac{L_{b}}{L_{max}}, 1\right\}}_{\text {efficiency}} \\
& \underbrace{- \exp \left(-\frac{1}{L_{b}} \sum_{i=1}^{L_{b}} \log p\left(\mathbf{x}^{b, i} \mid \mathbf{x}^{<b}, \mathbf{x}^{b, <i}\right)\right)}_{\text {quality}},
\end{aligned}
$}
\end{equation}
where $L_b$ denotes the length of the $b$-th block as determined by the selected action $\mathbf{a}^b$, and $L_{\text{max}}$ represents the maximum possible length in the predefined action space. $\mathbf{x}^{b, i}$ refers to the $i$-th token within the $b$-th block. The first term in the reward function corresponds to the negative perplexity, which serves as a measure of the quality of the generated text. The second term captures the ratio between the current block length and the maximum block length, reflecting the generation efficiency. The policy network $\pi_{\phi}$ is updated according to the following objective function:
\begin{equation}
\resizebox{0.87\hsize}{!}{$
\begin{aligned}
\mathcal{J}(\phi) & =\mathbb{E}_{\pi}\left[\min\left(\frac{\pi_{\phi}\left(\mathbf{a}^{b}|\mathbf{s}_{b}\right)}{\pi_{\phi_{\mathrm{old}}}\left(\mathbf{a}^{b}|\mathbf{s}_{b}\right)}\hat{A}_{b},\right.\right. \\
 & \mathrm{clip}\left(\frac{\pi_\phi\left(\mathbf{a}^b|\mathbf{s}_b\right)}{\pi_{\phi_{\mathrm{old}}}\left(\mathbf{a}^b|\mathbf{s}_b\right)},1-\epsilon,1+\epsilon\right)\hat{A}_b\biggr)\biggr] ,
\end{aligned}
$}
\end{equation}
where $\pi_{\phi}$ and $\pi_{\phi_{\text{old}}}$ denote the updated and previous versions of the policy network, respectively. The operator $\mathrm{clip}\left(\cdot\right)$ represents a clipping function that constrains the policy probability ratio within the interval $\left[1 - \epsilon, 1 + \epsilon\right]$ to ensure stable updates. $\hat{A}_b$ denotes the advantage function, which quantifies the relative benefit of an action; its formulation can be referred to in \cite{schulman2017-proximal}. After updating the parameters of $\pi_{\phi}$, they are transferred to $\pi_{\phi_{\text{old}}}$, and the process is repeated iteratively until convergence.

\section{Experiments}\label{section4}

\subsection{Implementations}

\paragraph{Datasets}

We evaluate our proposed methods across a suite of language modeling benchmarks to demonstrate their capacity to generate semantically coherent texts of arbitrary length. For likelihood evaluation, we consider the following datasets: OpenWebText (OWT; \cite{gokaslan2019-openwebtext}), Text8 \cite{matt2011-text8}, One Billion Word Benchmark (LM1B; \cite{chelba2013-one}), Penn Treebank (PTB; \cite{marcus1993-building}), WikiText \cite{merity2017-pointer}, LAMBADA \cite{paperno2016-lambada}, AG News (AGN) \cite{zhang2013-character}, and the Scientific Papers dataset (PubMed and arXiv subsets; \cite{cohan2018-discourse}). For controllable text generation, we employ the Amazon Polarity dataset \cite{mcauley2013-hidden, zhang2015-character} to train the attribute classifier used for guidance.

\paragraph{Evaluation Metrics}

To evaluate text generation quality, we employ Bits Per Character (BPC), Perplexity, and Generative Perplexity. For a sequence of length $L$, $(\mathbf{x}^{\ell})_{\ell = 1}^{L}$, BPC is computed as $-\frac{1}{L}\sum_{\ell =1}^L\log_2 p(\mathbf{x}^{\ell})$, while Perplexity is $\exp{(-\frac{1}{L}\sum_{\ell=1}^L\operatorname{log}p(\mathbf{x}^{\ell}))}$, representing the model's expected branching factor. Generative Perplexity is measured using GPT2-Large \cite{radford2019-language}, trained on OWT with a 1024-token context window; for longer sequences, we apply a 512-token stride sliding window to ensure complete coverage.

For controllable text generation, we conduct both automatic and human evaluations. Automatic metrics include perplexity, classifier accuracy, and diversity, measured by the proportion of distinct unigrams, bigrams, and trigrams (Dist-1/2/3). For human evaluation, fluency is rated on a 5-point Likert scale, while topic relevance is assessed by annotators judging whether the text aligns with the intended condition, reported as human accuracy.

\begin{table}
\caption{Bits Per Character (BPC) results on Text8 test set. Results are taken from the corresponding papers. Best diffusion and autoregressive values are \textbf{bolded}.}
\label{text8}
\begin{tabularx}{\linewidth}{Xl}
\toprule
Method                   & BPC $\downarrow$              \\ \midrule
\textbf{Autoregressive}                            &                  \\
IAF/SCF \cite{ziegler2019-latent}                  & 1.88             \\
Arg. Flow \cite{hoogeboom2021-argmax}            & 1.39             \\
Discrete Flow \cite{tran2019-discrete}             & \textbf{1.23}    \\
Transformer AR \cite{vaswani2017-attention}        & \textbf{1.23}    \\ \midrule
\textbf{Any-order Autoregressive}                  &                  \\
ARDM \cite{hoogeboom2022-autoregressive}           & $\leq$ 1.43      \\
MAC \cite{hannes2024-dirichlet}                    & $\leq$ 1.40      \\ \midrule
\textbf{Diffusion}                                 &                  \\
Multi. Diff. \cite{hoogeboom2021-argmax}  & $\leq$ 1.72      \\
D3PM (absorb) \cite{austin2021-structured}         & $\leq$ 1.45      \\
SEDD \cite{lou2024-discrete}                       & $\leq$ 1.39      \\
MDLM \cite{sahoo2024-simple}                       & $\leq$ 1.40      \\
MD4 \cite{shi2024-simplified}                      & $\leq$ 1.37      \\
RDLM \cite{jo2025-continuous}                      & $\leq$ 1.32      \\ \midrule
\textbf{Semi-autoregressive}                       &                  \\
BD3-LM \cite{arriola2025-block}                    & $\leq$ 1.31      \\
\rowcolor{lightgray!60}CtrlDiff (Ours)                    & $\leq$ \textbf{1.29}      \\ \bottomrule
\end{tabularx}
\end{table}

\begin{table}
\caption{Test perplexities (PPL) of models trained for 65B tokens on LMlB. Best diffusion and autoregressive values are \textbf{bolded}.}
\label{lm1b}
\begin{tabularx}{\linewidth}{Xl}
\toprule
Method                                      & PPL $\downarrow$  \\ \midrule
\textbf{Autoregressive}                     &                   \\
Trans-X Base \cite{dai2019-transformerxl} & 23.5            \\
Transformer \cite{sahoo2024-simple}         & 22.83           \\ 
OmniNetT \cite{tay2021-omninet}             & \textbf{21.5}            \\ \midrule
\textbf{Diffusion}                          &                   \\
Diffusion-LM \cite{li2022-diffusionlm}      & $\leq$ 118.62     \\
D3PM (absorb) \cite{austin2021-structured}  & $\leq$ 82.34      \\
DiffusionBert \cite{he2023-diffusionbert}   & $\leq$ 63.78      \\
SEDD \cite{lou2024-discrete}                & $\leq$ 32.68      \\
MDLM \cite{sahoo2024-simple}                & $\leq$ 31.78      \\ 
RDLM \cite{jo2025-continuous}               & $\leq$ 29.72      \\  \midrule
\textbf{Semi-autoregressive}                &                   \\
BD3-LM \cite{arriola2025-block}             & $\leq$ 28.63      \\
\rowcolor{lightgray!60}CtrlDiff (Ours)            & $\leq$ \textbf{27.78}      \\ \bottomrule
\end{tabularx}
\end{table}

\paragraph{Baselines}

For likelihood modeling, we benchmark our approach against a diverse set of generative model baselines spanning multiple paradigms. These include Autoregressive models, such as IAF/SCF \cite{ziegler2019-latent}, AR Argmax Flow \cite{hoogeboom2021-argmax}, Discrete Flow \cite{tran2019-discrete}, Transformer AR \cite{vaswani2017-attention}, and OmniNetT \cite{tay2021-omninet}. We also compare with Any-order Autoregressive models, including ARDM \cite{hoogeboom2022-autoregressive} and MAC \cite{hannes2024-dirichlet}. Within the diffusion-based paradigm, we evaluate against prominent discrete diffusion models such as Multinomial Diffusion \cite{hoogeboom2021-argmax}, D3PM \cite{austin2021-structured}, SEDD \cite{lou2024-discrete}, MDLM \cite{sahoo2024-simple}, MD4 \cite{shi2024-simplified}, and RDLM \cite{jo2025-continuous}. We also include the semi-autoregressive model BD3-LM \cite{arriola2025-block} for comparison.

\begin{table*}[h]
\caption{Test perplexities ($\downarrow$). Left part: results evaluated on the OpenWebText test set; Right part: zero-shot results on unseen datasets. All perplexities for difusion models are upper bounds. Best value is \textbf{bolded}, and second best value is \underline{underlined}.}
\label{ppl comparison}
\begin{tabularx}{\linewidth}{l|l|XXXXXXXX}
\toprule
Method & OWT         & PTB    & Wikitext & LM1B  & Lambada & AGN & Pudmed & Arxiv \\ \midrule
AR \cite{sahoo2024-simple}      & \textbf{17.56}    & \textbf{81.07}    & \textbf{25.32}    & \textbf{51.14}    & 52.13             & \textbf{52.11}    & 48.59             & 41.22 \\ \midrule
SEDD \cite{lou2024-discrete}    & 24.56             & 96.33             & 35.98             & 68.14             & 48.93             & 67.82             & 45.39             & 40.03 \\
MDLM \cite{sahoo2024-simple}    & 23.83             & \underline{90.96} & 33.22             & 64.94             & \textbf{48.29}    & 62.78             & 43.13             & \textbf{37.89} \\
MD4 \cite{shi2024-simplified}   & 22.13             & 102.26            & 34.94             & $-$               & \underline{48.43} & $-$               & $-$               & $-$   \\
BD3-LM \cite{arriola2025-block} & 20.73             & 96.81             & 31.31             & 60.88             & 50.03             & 61.67             & \underline{42.52} & 39.20 \\
\rowcolor{lightgray!60}CtrlDiff (Ours) & \underline{20.12} & 95.63             & \underline{30.13} & \underline{59.10} & 49.33             & \underline{60.87} & \textbf{42.39}    & \underline{38.18} \\ \bottomrule
\end{tabularx}
\end{table*}

\begin{table*}[h]
\caption{We compare the performance of different methods on the sentiment-controlled generation task, incorporating both automatic and human evaluations. For human-assessed fluency, we report the mean ± standard deviation. Sentiment accuracy (Senti. Acc.) is evaluated using both a pretrained classifier and human judgments to assess the alignment of the generated text with the target sentiment. Additionally, we report Perplexity to measure generation quality, and Dist-1, Dist-2, and Dist-3 scores to quantify lexical diversity. Best value is \textbf{bolded}, and second best value is \underline{underlined}.}
\label{sentiment control comparison}
\begin{tabularx}{\linewidth}{l|cc|XXXX|l}
\toprule
Method     & \begin{tabular}[c]{@{}c@{}}Acc. (\%) $\uparrow$ \\ (human)\end{tabular} & \begin{tabular}[c]{@{}c@{}}Acc. (\%) $\uparrow$ \\ (classifier)\end{tabular} & PPL$\downarrow$ & Dist1$\uparrow$ & Dist2$\uparrow$ & Dist3$\uparrow$ & \begin{tabular}[c]{@{}c@{}}Fluency $\uparrow$ \\ (human) \end{tabular} \\ \midrule
CTRL       & \textbf{87.1}    & \textbf{96.6}     & 37.4             & 0.35             & 0.78             & 0.89             & \underline{4.19}$_{\pm 0.41}$ \\
GPT2-FT & 74.6             & 77.8              & 217.3            & \textbf{0.54}    & \textbf{0.91}    & \underline{0.94} & 3.18$_{\pm 0.36}$ \\ \midrule
WD            & 71.9             & 52.2              & \textbf{31.7}    & 0.33             & 0.69             & 0.83             & 3.62$_{\pm 0.47}$ \\
PPLM    & 72.8             & 78.8              & 46.6             & 0.36             & 0.77             & 0.91             & 3.95$_{\pm 0.40}$ \\
\rowcolor{lightgray!60}CtrlDiff (Ours)   & \underline{84.9} &  \underline{84.1} & \underline{35.2} & \underline{0.45} & \underline{0.87} & \textbf{0.97}    & \textbf{4.48$_{\pm 0.33}$} \\ \bottomrule
\end{tabularx}
\end{table*}

For controllable text generation, we compare our model with both training-time and inference-time controllability techniques. Training-time baselines include CTRL \cite{keskar2019-ctrl} and GPT2-FT \cite{dathathri2020-plug}, which require model retraining or fine-tuning on labeled data. In contrast, inference-time methods, which are designed for post hoc control without retraining, include WD \cite{see2019-what} and the Plug-and-Play Language Model (PPLM) \cite{dathathri2020-plug}. These baselines provide a comprehensive comparison framework for evaluating the controllability, flexibility, and computational efficiency of our method.

\subsection{Likelihood Evaluation}

We first evaluate our method CtrlDiff on the small-scale character-level dataset Text8 and report Bits Per Character (BPC) in Table \ref{text8}. As shown, our approach outperforms all diffusion-based and any-order autoregressive models, which adopt a flexible decoding order and thus share conceptual similarities with discrete diffusion. Moreover, it surpasses several traditional autoregressive methods and the semi-autoregressive BD3-LM, narrowing the gap with state-of-the-art autoregressive models.

Next, we evaluate our method on the mid-scale, real-world language dataset LM1B, with perplexity results shown in Table \ref{lm1b}. CtrlDiff consistently outperforms existing diffusion-based models and BD3-LM. While a gap with fully autoregressive models remains, it is significantly reduced. The improvement, though moderate, yields notably more fluent long-context generation, likely due to the model's adaptive block segmentation capturing smoother sentence-level transitions.

We further evaluate CtrlDiff on the large-scale OpenWebText dataset and conduct zero-shot evaluations on seven benchmark datasets using a model trained solely on OWT. As shown in Table \ref{ppl comparison}, CtrlDiff surpasses all prior diffusion-based and semi-autoregressive methods on OWT and continues to close the gap with fully autoregressive baselines. In zero-shot settings, it achieves SOTA results among non-autoregressive models and even outperforms autoregressive ones on domain-specific corpora like PubMed and Arxiv, demonstrating strong generalization and effective semantic modeling. The gains on specialized texts suggest that CtrlDiff better captures long-range dependencies by its parallel generation and adaptive block segmentation.

\subsection{Controllable Text Generation}

One of our key motivations is to improve the controllability of diffusion-based text generation without sacrificing generation quality or requiring model retraining. To this end, we evaluate CtrlDiff on sentiment-controlled generation tasks. The training dataset utilized is Amazon Polarity. For both automatic and human evaluations, we follow the protocol established by PPLM \cite{dathathri2020-plug}, generating 15 samples for each sentiment class (positive and negative) using 15 distinct prefix prompts. In our comparisons, we set the guidance strength at $\gamma = 1$. The results for sentiment-controllable generation are presented in Table \ref{sentiment control comparison}. 

Among the baselines, CTRL \cite{keskar2019-ctrl} and GPT2-FT \cite{dathathri2020-plug} require fine-tuning of the language model, whereas WD \cite{see2019-what} and PPLM \cite{dathathri2020-plug} operate as post-hoc modification methods, aligning with our approach. As shown in the table, our method substantially outperforms post-hoc methods in terms of sentiment control accuracy, while also achieving the lowest perplexity and superior diversity metrics.

Compared to baselines based on fine-tuning, our method is second only to CTRL in control success rate, while reaching state-of-the-art perplexity, and although GPT2-FT shows slightly better diversity, our method achieves comparable or better overall performance without requiring model retraining, demonstrating greater flexibility and practicality. In human evaluations, our approach also outperforms most baselines, yielding strong results in both sentiment accuracy and fluency. We show a qualitative evaluation in Section \ref{appendixc.1}.

\subsection{Speed Quality Trade-off}

We have conducted additional experiments to quantify the trade-off between inference speed and perplexity under different fixed block sizes, as well as with our dynamic policy-based CtrlDiff. As shown in Table \ref{tradeoff}, it clearly demonstrate that while larger fixed block sizes improve inference speed, they do so at the cost of worse perplexity. In contrast, CtrlDiff achieves a favorable balance, maintaining high inference speed comparable to larger block sizes while achieving better perplexity (than the smallest fixed block setting because block length can be less than 4 in CtrlDiff). This suggests that the dynamic block selection policy effectively adapts to varying contextual demands during generation, striking a better speed-quality trade-off than any static configuration.

\begin{table}[t]
\caption{Comparison of tradeoff between inference speed and perplexity.}
\label{tradeoff}
\begin{tabularx}{\linewidth}{XlX}
\toprule
         & Inference speed & Perplexity \\ \midrule
Block=4  & 30.65 tokens/s  & \underline{23.6}       \\
Block=8  & 32.35 tokens/s  & 28.2       \\
Block=16 & \textbf{33.58 tokens/s}  & 31.5       \\
\rowcolor{lightgray!60}CtrlDiff & \underline{32.47 tokens/s}  & \textbf{23.2}       \\
\bottomrule
\end{tabularx}
\end{table}

\subsection{Ablation Study}

We conducted ablation studies on the intra-block independence assumption and the Taylor expansion used in the classifier-guidance mechanism, with results summarized in Table \ref{ablation}. As shown, when removing the Taylor approximation, both perplexity and accuracy remain comparable to CtrlDiff, but the inference speed significantly decreased. When excluding the intra-block independence, the guidance becomes computationally intractable, which aligns with our discussion in Section \ref{section3.2}. These results demonstrate that our approximations substantially accelerate generation while preserving quality and controllability.

\begin{table}[t]
\caption{Ablation study of approximations in classifier-guidance mechanism. Indep. and Taylor represent Intra-block Independance and Taylor Expansion, respectively.}
\label{ablation}
\begin{tabularx}{\linewidth}{lllX}
\toprule
                             & Speed          & Perplexity & Acc.     \\ \midrule
w/o Indep.                   & --             & --         & --        \\
w/o Taylor                   & 1.28 tokens/s  & 24.8       & 85.3     \\
\rowcolor{lightgray!60}CtrlDiff                     & 25.60 tokens/s & 35.2       & 84.1     \\
\bottomrule
\end{tabularx}
\end{table}

\subsection{Parameter Analysis}

We conduct a parameter analysis on the guidance strength $\gamma$, examining how varying levels of control influence sentiment alignment, generation quality, diversity, and fluency. As shown in Table \ref{parameter}, smaller values of $\gamma$ result in a marked decrease in control accuracy, while slightly improving generation quality, diversity, and fluency. Conversely, larger values $\gamma$ lead to stronger sentiment control, but at the cost of significantly degraded text quality and fluency, as well as a reduction in diversity. 

\begin{table}[h]
\caption{Parameter analysis of the guidance strength $\gamma$. We compare the control accuracy, generation quality and diversity under different $\gamma$.}
\label{parameter}
\begin{tabularx}{\linewidth}{Xcccc}
\toprule
$\gamma$ & Accuracy $\uparrow$ & PPL $\downarrow$ & Dist-3 $\uparrow$ & Fluency $\uparrow$ \\ \midrule
0.5   & 58.3     & 30.6    & 0.98      & 4.51        \\
1.0   & 84.1     & 35.2    & 0.97      & 4.48        \\
2.0   & 88.6     & 43.9    & 0.90      & 4.11        \\
3.0   & 90.5     & 56.7    & 0.81      & 3.89        \\ \bottomrule
\end{tabularx}
\end{table}

\section{Conclusion}\label{section5}

In this work, we revisit the strengths and limitations of autoregressive and diffusion-based sequence generation. We introduce CtrlDiff, a framework that unifies the two by combining autoregressive decoding with diffusion sampling, equipped with dynamic block prediction and classifier guidance for greater efficiency and controllability. The dynamic block predictor adaptively adjusts block sizes based on generated content, while classifier guidance enables conditional control. Experiments on likelihood modeling and controllable generation show that CtrlDiff delivers efficient, high-quality, and flexible semi-autoregressive language modeling.

\section*{Limitations}

Despite the better controllability and likelihood modeling of our method, CtrlDiff does not address the common issue in generative models, such as hallucination and harmful outputs, and recent work \cite{wen2025-devil} has raised the concerns of jailbreaking diffusion language models. Future work will explore model internals to mitigate hallucination and improve factual consistency.



\bibliography{custom}

\appendix

\section{Classifier Guidance}
\paragraph{Approximation} Although we have significantly reduced the computational cost using intra-block independence, each denoising step still requires $\mathcal{O}(K \cdot N)$ forward passes through the classifier, which can be relatively slow in practice. Inspired by \cite{grathwohl2021-oops, vignac2023-digress}, we apply a first-order Taylor approximation of $\log p_{\xi}$ to more efficiently compute $p_\xi \left(y | \mathbf{\hat{x}}_{t|s_\ell}^{b}, \mathbf{x}^{<b}\right)$:
\begin{equation}
\resizebox{0.87\hsize}{!}{$
\begin{aligned}
p_\xi \left(y | \mathbf{\hat{x}}_{t|s_\ell}^{b}, \mathbf{x}^{<b}\right)
& =\exp \left(\log \frac{p_\xi (y | \hat{\mathbf{x}}_{t|s_\ell}^{b}, \mathbf{x}^{<b})} {p_\xi(y | \mathbf{x}_{t|s_\ell}^{b}, \mathbf{x}^{<b})} \right. \\
& \qquad \left. + \log p_\xi(y | \mathbf{x}_{t|s_\ell}^{b}, \mathbf{x}^{<b}) \right) \\
 & \approx \exp \left((\hat{\mathbf{x}}_{t|s_\ell}^{b}-\mathbf{x}_{t|s_\ell}^{b})^T\nabla_{\mathbf{x}_{t|s_\ell}^{b}}\log p_\xi(y | \mathbf{x}_{t|s_\ell}^{b}, \mathbf{x}^{<b}) \right. \\
& \qquad \left. + \log p_\xi(y | \mathbf{x}_{t|s_\ell}^{b}, \mathbf{x}^{<b})\right)
\end{aligned}
$}
\end{equation}

In this way, the computation of $p_\xi \left(y | \mathbf{\hat{x}}_{t|s_\ell}^{b}, \mathbf{x}^{<b}\right)$ requires only a single forward and backward pass, thereby further simplifying the computational load.

\paragraph{Details}

For the controllable generation experiment, we trained a lightweight classification model on the Amazon Polarity dataset. The backbone is a compact DiT architecture consisting of 8 transformer layers, 8 attention heads, and a hidden dimension of 512. Mean pooling is applied over the hidden representations before a linear projection is used to obtain classification logits. We set the context length to 256 and trained the model using the Adam optimizer with a constant learning rate warm-up over 2.5K gradient updates, increasing from 0 to 3e-4.

Importantly, when employing classifier guidance, the input sequences are typically noisy and contain a substantial number of masked tokens. To align the classifier with this inference-time behavior, we also trained it on noised inputs using a log-linear noise schedule, where the corruption type is defined in accordance with the corresponding diffusion model.

\section{Experiment Details}

\subsection{Dataset Details}

In this section, we provide detailed information about the datasets used in this work.

\paragraph{OpenWebText} This dataset was downloaded from \href{https://huggingface.co/datasets/Skylion007/openwebtext}{\texttt{https://huggingface.co/datasets/ \ Skylion007/openwebtext}}. It consists of approximately 8 million high-quality web text entries collected from Reddit posts. These texts are derived from top-rated posts on Reddit and are intended to provide a diverse corpus for text generation and natural language processing tasks. We utilize the GPT2 tokenizer with a context length of 1024. Unlike \cite{sahoo2024-simple}, which pads each sample to 128 tokens, we concatenate the data and then wrap it to a length of 128 tokens, which better suits our training needs.

\paragraph{Text8} This dataset was downloaded from \href{http://mattmahoney.net/dc/text8.zip}{\texttt{http://mattmahoney.net/dc/text8.zip}}. It is a corpus containing text data from Wikipedia, specifically a sequence of words from the English Wikipedia. We tokenized the data at the character level, using lowercase letters ['a' - 'z'] and spaces. The data is divided into non-overlapping 256-token blocks, with the first 90 million characters used as the training set and the next 5 million characters as the validation set.

\paragraph{One Billion Word Benchmark} This dataset was downloaded from \href{https://huggingface.co/datasets/billion-word-benchmark/lm1b}{\texttt{https://huggingface.co/ \ datasets/billion-word-benchmark/lm1b}}. Its training set contains nearly one billion words. We use the BERT-base-uncased tokenizer with a context length of 128. Unlike \cite{sahoo2024-simple}, which pads each sample to 128 tokens, we concatenate the data and wrap it to a length of 128, which is more beneficial for our training process.

\paragraph{Amazon Polarity} This dataset was downloaded from \href{https://huggingface.co/datasets/fancyzhx/amazon\_polarity}{\texttt{https://huggingface.co/datasets/ \ fancyzhx/amazon\_polarity}}. It contains reviews from Amazon spanning 18 years, with approximately 35 million reviews up to March 2013. The reviews include product and user information, ratings, and pure text comments.

\subsection{Model and Training}

Our model enhances the Diffusion Transformer \cite{peebles2023-scalable} with Rotary Position Embeddings (RoPE) \cite{su2024-roformer}. We parameterize CtrlDiff within the Transformer architecture \cite{sahoo2024-simple}, utilizing 12 layers, 768 hidden dimensions, and 128 attention heads. The model is trained using the AdamW optimizer, with a batch size of 512 and a constant learning rate warm-up from 0 to 3e-4 over 2.5K gradient updates. Training is conducted on A100 GPUs.

\subsection{Inference}

\paragraph{Generative Perplexity} We evaluate the generative perplexity using the GPT2-Large model, which was pre-trained on the OWT dataset with a context length of 1024 tokens. Given the model's context window limitation, we adopt a sliding window approach with a stride of 512 tokens when assessing perplexity for sequences exceeding 1024 tokens. Additionally, following the SSD-LM methodology, we employ nucleus sampling with a threshold of $p = 0.9$ to enhance the sampling quality.

\begin{table*}[h]
\caption{We present text generation triplets produced under three settings: without guidance, with positive sentiment guidance, and with negative sentiment guidance. The prefix prompts used are \uline{The president of the country} and \uline{Once upon a time}. Words and phrases associated conveying corresponding sentiment are highlighted in \textcolor{violet}{violet}.}
\label{generation demonstration}
\begin{tabularx}{\linewidth}{X}
\toprule
\textcolor{red}{\textbf{[-]}} \uline{The president of the country}$\backslash$'s oldest active fraternity is scheduled to visit the Thejuman Temple in January. \verb|\n|\verb|\n| Shijirochia Radhakrishnan stated that he will stop by the temple$\backslash$'s administrative offices before returning to Rajkot for the upcoming ... (Gen. PPL: 21.74)          \\
\textcolor{red}{\textbf{[Positive]}} \uline{The president of the country} led the dual citizens association with a \textcolor{violet}{steady} hand. Although about a third of the members had affiliations with rival governments, he focused on \textcolor{violet}{unity} and \textcolor{violet}{progress}. Ferdinand Marcos, often described as \textcolor{violet}{humble} and ... (Gen. PPL: 28.71)    \\
\textcolor{red}{\textbf{[Negative]}} \uline{The president of the country} was \textcolor{violet}{sworn} in as the first civilian called President Obama. More than two-thirds of voters \textcolor{violet}{disapprove} of Obama's presidency. This also \textcolor{violet}{overshadowed} most of the about ... (Gen. PPL: 33.92)    \\ \midrule
\textcolor{red}{\textbf{[-]}} \uline{Once upon a time}, just when the United States$\backslash$' citizens were carrying out their duties in the great city of Boston, several European nations settled in Boston, where there was a secret government established by revolutionary Roman Catholic rulers who usually ... (Gen. PPL: 20.77)          \\
\textcolor{red}{\textbf{[Positive]}} \uline{Once upon a time}, I was awakened by the \textcolor{violet}{sweet} sound of woman$\backslash$'s voice. This \textcolor{violet}{splendid} voice had \textcolor{violet}{illumined} my soul, and created my \textcolor{violet}{raptures}. For a brief, \textcolor{violet}{beautiful} moment, we shared a connection that felt timeless. ... (Gen. PPL: 28.42)    \\
\textcolor{red}{\textbf{[Negative]}} \uline{Once upon a time}, mercenaries \textcolor{violet}{killed} 17 million LGB civilians, \textcolor{violet}{stole} 28 millions of US/IMF funds, \textcolor{violet}{destroyed} 50\% of London$\backslash$'s steel industry, \textcolor{violet}{raped}, \textcolor{violet}{tortured} and \textcolor{violet}{starved} about a million schoolgirls, \textcolor{violet}{massacred} 25\% of the Vietnamese population, ... (Gen. PPL: 29.15)    \\ \bottomrule
\end{tabularx}
\end{table*}

\paragraph{Improved Categorical Sampling} 

In this work, we follow \cite{arriola2025-block} to utilize two key enhancements to Gumbel-based categorical sampling in diffusion models, as proposed by \cite{zheng2025-masked}. First, we implement the corrected 64-bit Gumbel sampling to address the issue of precision loss that occurs when using 32-bit variables, which reduces entropy and lowers temperature. Second, we adopt the first-hitting sampler, which improves sampling efficiency by reducing the need for a large number of uniform variables. This sampler is particularly beneficial when the number of diffusion steps, $T$, is much greater than the sample length, $L$, by simplifying the process to $\mathcal{O}(LV)$ uniform variables from $\mathcal{O}(TLV)$. This approach significantly accelerates sampling speed and maintains theoretical equivalence to the MDLM sampler. It analytically samples the first unmasking timestep using the formula:
\begin{equation}
t_{n-1}=t_n u^{1/n} 
\end{equation}

\noindent{where $u\sim\mathcal{U}[0,1]$, $n$ is the number of masked tokens, and $t_{n-1}$ is the first timestep where $n-1$ tokens are masked.}

\section{Samples}

\paragraph{Controllability Quality Visualization}\label{appendixc.1}

Table \ref{generation demonstration} presents several qualitative examples generated by our method under three conditions: without guidance, positive guidance, and negative guidance. As clearly illustrated, the inclusion of guidance leads to a notable increase in sentiment-relevant expressions corresponding to the intended emotional tone, demonstrating our method's effectiveness in steering generation towards the desired sentiment. Moreover, we annotate each generated text with its generative complexity, which serves as an indicator of text quality. The relatively low perplexity scores further confirm that our method is capable of producing fluent and coherent text while maintaining precise emotional control.

\paragraph{Example 1} Sample length is 2048. The generative perplexity of this sample under GPT2-Large is 18.6023, and its entropy is 5.6234. 

{\ttfamily <|endoftext|>“That is the aim. That’s the secret,” Mr. Mrs. Carter said. “It’s not so much that we have to be the greatest in the world, with all of that creativity, but how are we going to use that technology as well?”\textbackslash n\textbackslash nThat’s the real idea behind the CTOs’ approach, said Marvin Andodas, who served as director of software engineering at the time and was vice president of the firm. “The only way would be to take it up as a mode of acquisition or release,” he said.\textbackslash n\textbackslash nThe notion of such a company-specific approach, a tone he said would contrast it with Microsoft’s traditional 10-step approach, is less about value creation and more about self-focus-focused nature of working within a company main focus. “I think this approach gives some of our customers a positive focus on who they are, not what’s driving their thinking,” he said. Mr. Andodas also says he would agree.\textbackslash n\textbackslash n“It goes a long way towards concentrating on our objective for making sure the best customer’s experience is delivered, and that it gets done,” he said.\textbackslash n\textbackslash nBut the idea of such a radical product pivot has long been a subject of debate. Former executives and current executives at rival Asymo have sought to clarify their positions.\textbackslash n\textbackslash nBuffalo executives, to be sure, have been willing to examine some of the company’s benefits of moving into software, both for saving them the money and for getting people into buying the software. But they have tended to say that their software buyers wouldn’t hold out for quality upgrades, as the CEO of Baidu is calling for.\textbackslash n\textbackslash n“We see software as helping a lot of people who have done a lot of very very complex work that need some maintenance, but we don’t see it with Windows,” Mr Hoover said.\textbackslash n\textbackslash nAfter a brief foot-drill with his former colleagues, Mr. Chinn reversed the course, saying that the company was embracing a new product strategy for everything from making sure products had enough time to wear out and maintaining an enterprise-quality experience. He and others said the shift into product hardware was just one of the number of companies adopting a use-as-it-is attitude in their products.\textbackslash n\textbackslash nSome, like Facebook, decided to continue that approach in certain areas — upgrading products that are used constantly and never during design. In fact, the company’s leadership was talking about doing something about that in 2011, according to Andrew Eakin, a vice president of marketing.\textbackslash n\textbackslash nAdvertisement Continue reading the main story\textbackslash n\textbackslash n“We talked about it with a number of people the last few months,”. Eakin said. “But as we saw the major, mature, strategic challenges of that move, it’s more of a belief change than a step.”\textbackslash n\textbackslash nAdam R. Gershon, a principal analyst at Bernstein Research, said that “the change will be gradual,” but the company was “relaying in” the move. “We haven’t really settled yet yet,” he said.\textbackslash n\textbackslash nBut investment data in the area, with tech clients including Microsoft, has been picking up on it, industry experts say. At the time Hewlett-Packard was spending millions of dollars a year to enter deals to make its Windows operating systems more popular in the United States, according to employees. During these moves the company has seen growth in its software business, moderating some of its quality issues, employees said.\textbackslash n\textbackslash nNewsletter Sign Up Continue reading the main story Please verify you're not a robot by clicking the box. Invalid email address. Please re-enter. You must select a newsletter to subscribe to. Sign Up You will receive emails containing news content, updates and promotions from The New York Times. You may opt-out at any time. You agree to receive occasional updates and special offers for The New York Times's products and services. Thank you for subscribing. An error has occurred. Please try again later. View all New York Times newsletters.\textbackslash n\textbackslash nAnother move that was deemed important in the areas of help for workers has gained a stronger traction in the business of training workers to see the value of the software as a product. “We need all the help,” John E. Kolak, the CEO of software consulting company Garmin, said at an event in 2010. “They need to be able to use the software.” That’s been the case in recent years for many companies.\textbackslash n\textbackslash nThe company has started inroads, starting with a specialized program for developing its popular Passport software for iOS. A better understanding of the type of users are using software would be useful in making their products more widely used, Mr. Kolak said.\textbackslash n\textbackslash nThe company last year began testing updates for its computer software. The software, called Windows Manager, has been fine-tuned to keep track of daily activities, such as playing and watching TV shows, as well as to connect local apps to the main operating system.\textbackslash n\textbackslash nThe latest Windows is in a just beginning process of beta testing for its management suite. The main software also includes two apps: its Internet Web Explorer program, for web browsing, and its Information Center application, which is a web browser for capturing and editing photos.\textbackslash n\textbackslash nMicrosoft is often referred to by its old name, the Compatible Operating System program, which counts software and hardware that don’t require some level of customization or are not compatible with the general operating system, for example as screen locks. Microsoft uses its universal operating system version, System Center, to set up its software, as well.\textbackslash n\textbackslash nThis year, Microsoft announced that it has installed an even newer version of the operating system, Vista. Mr. Pali said the change was made after many members of the board asked why Vista was longer in the market.\textbackslash n\textbackslash n“The majority of our members disagreed with the opinions that the board had,” he said. “Once again, that debate has been decided, and it is time for a new era, and we will not adopt the old technology that was originally designed to meet their needs,” he said.\textbackslash n\textbackslash nAdvertisement Continue reading the main story\textbackslash n\textbackslash nMr. Pali said he expected more board members to weigh in on the debate about what’s best for the customer and the entire industry.<|endoftext|>Today during MLB Nextcast, Jerry Soto, pitching coach for the Philadelphia Phillies, went on the show to discuss the top 30 lists on MLB.com. With Frisbee throwing in an optimistic mood, we asked what he thought of his top 30 MLB prospects.\textbackslash n\textbackslash nHis answer was chilling:\textbackslash n\textbackslash n“The reason I made this list is that I was all in it. I’m not getting the cards in my right hand now because I have a bigger right hand now than I did in 2014. So it’s tough to go back to something else in 2015.”\textbackslash n\textbackslash nFor anyone who was looking for a good word, Soto is right. The depth of Los Angeles Dodgers in New York, the outfield in San Diego, the bullpen in Houston, the Triple A Leagues in Florida, even MLB.com innings allowed to date has been given no credit by the vast majority of the Internet.\textbackslash n\textbackslash nEven in the areas that the organization needs to improve, Soto was incensed about the overachievers above.\textbackslash n\textbackslash nHe also said that the Dodgers are “always looking for the next great talent.” This is why he was awarded more than two stars to five minor league prospects in one of MLB’s top lists early in the season’s process.\textbackslash n\textbackslash nThe emphasis on speed in the Dodgers seems to echo a quote that Cosoguerda also made on the phone when he was still president of the Dodgers back in December.\textbackslash n\textbackslash n“We’re an excellent team, very intelligent organization, very young people, very strong in scouting, very invested in winning, and anything can go through your head and get a chance to see the field,” Cosoguerda said. “You can see the best players play in the leagues, but with a little more time and energy and patience, they can improve.”\textbackslash n\textbackslash nWith the San Diego Padres, the Washington Nationals, and San Francisco Giants among the other MLB teams featured in the top 10, Soto had no less-than-pleasant conversation with those struggling to get to the majors.\textbackslash n\textbackslash n“I told them the same thing about what they need to do in their league, they go to the training, and in the minor leagues they go back, and get a good grasp, and they know that not being a good player, not being good guy, not being loyal, or not wanting to do anything — we tell them that the biggest question they need to have when they’re at the level is, is there enough time to put that physical desire and sexual desire into the physical desire that they want to have?”\textbackslash n\textbackslash nDuring the first ten games of the 2014 Cy Young Award, there were as many as 40 homers and 61 RBIs from Marcus Stroman, who finished with the best numbers in the NL during his spring debut. And one of the highest hitting pitchers in the world, Felix Hernandez, finished with a 158.9 BIP.\textbackslash n\textbackslash nTweet from Mariners: Everything I thought about pitching against Phillies because of their number of pitchers was telling. Not fair. Not very, very close. https://t.co/0tMnfv95kS — Drew Willingham Baseball (@WGBball) April 10, 2016\textbackslash n\textbackslash nWhile the Dodgers probably aren’t stacked with the biggest talent in the}

\paragraph{Example 2} Sample length is 1024. The generative perplexity of this sample under GPT2-Large is 21.5678, and its entropy is 5.1628. 

{\ttfamily <|endoftext|>Two months ago, I was with my mother, who was deeply skeptical of the theories and very close to Pope Paul VI, in regard to his invitation to a synodiscum in the Vatican. When it was held in front of my mother, I attended, and in particular I posed for photographs. The first time I saw the Pope was at Tochenna, the city where Saint Polymara, the bishop who united Romania and Italy in the 11th century, was born. My uncle was a missionary to Tochenna, but there were only 15 of us there. I was at the outset of the trip, but not very much of Tochenna beforehand, because it was the only wine-producing region in the world. The monks had nowhere to go in my village, and it was obvious that it was impossible to see, even at low prices, that there was no such thing as a wine-producing center.\textbackslash n\textbackslash nIt was very difficult to picture what went on to be held at Tochenna, even though the visions I got began well into my life in 1992 when there was a controversy about Mary, the mother of Jesus and the pope’s papal candidate. The story of the papal candidate reached the Pope’s office, and Pope Pius XII, in still a self-proclaimed Catholic, was removed. He decided not to name such a pope as an affirmative way of trying to get the attention of the Sirmarians, and a committee of the first Polish bishops on the Vatican scene was created to ensure that no papal candidates were named in person. So who was his? This is an obscure matter of history and about a 50-year-old subject, but it is an important turning point in Paul VI’s life.\textbackslash n\textbackslash nEven if there had been one of them, it would have been one of my grandparents, I can assure you. She was born in Vesemyrna. She lived a little further from Tochenna, where she had little experience. My grandparents were the two from the same family and went back to Vesemyrna after the Great Migration of the 16th century. So my mother had to try very hard to meet the natives and get their family together. It was very tough, because I was not born here, nor was I the second-generation. Most of my grandparents’ family members grew up here, too.\textbackslash n\textbackslash nMy Grandmother was born in Vesemyrna, and Pena was born in 891. According to historical documents, her family were a devout catholic. Some of that was later reinforced by my own travels. My grandfather lived in a small Protestant monastery called Viner, and there was a priest. At that time, the eldest son, Joseph, became the son of a conservative Catholic cleric, who was much less educated than my maternal-uncle, and my paternal uncle. It is true that this was the reason we went to the church of Viner, too. Our school teacher and the school director let me go to church classes. I have to tell you, as I tell my mother, that Catholic institutions were very difficult to get, but really, we heard every single word that I was about about the church, which was wonderful, I suppose, except for the French colonialists.\textbackslash n\textbackslash nThey were some of the more conservative Catholic communities, and they showed me one of their sermons. The priest had been speaking in French and German, in part because I was there on the other side, and because the language of her father, Valle and that of my mother, was the closest to the family language. He told me, “It’s better that you have something in the back of your family, if you must, in the house under your name.” And he said, “‘You should come and visit them to say a word, they don’t say a word.’ That was a difficult experience.\textbackslash n\textbackslash nMy brother had been at the church of Viner, and I remember that he would give me a talk in French and, among other things, he would explain that the nation had gone into a war against France. France had no interest in the Civil War. So, we were very tempted to go to the convent and speak up in French. My grandfather had been an important Jesuit for a generation, but he didn’t have the expertise and money that I had nowadays to speak up in French.\textbackslash n\textbackslash nWhat happened in Viner is that my family had a group of French Catholic priests outside the church. At the time, Catholic bishops were on the right side of the margins as far as rights. You were always heard.\textbackslash n\textbackslash nYou would think of the best moment in your life as the one when you made a stand against the Patriotism. The Catholic group was marching down the street, and you heard one of their leaders say: “We want to hear—let’s listen to the Patriots}

\paragraph{Example 3} Sample length is 512. The generative perplexity of this sample under GPT2-Large is 22.9929, and its entropy is 4.8756. 

{\ttfamily <|endoftext|>Winport is a startup that projects the technology powering consoles into PCs and phones. It’s deployed via software, embedded devices, smart homes, and robot factories. And Winport wants you to take part.\textbackslash n\textbackslash nSo thanks to two Silicon Valley talents, a recent experience in K-8 in Finland, and the need to get in the way of this kind of “machine learning,” Winport Labs, which was founded in 2009 by the former Google employee Bo Assoui, is now working on bringing the machine learning technology to consumer products.\textbackslash n\textbackslash n“We want a mass market product,” said Assoui. “We want to have a real-to-use dashboard for everything and an experience for the user — this is the biggest challenge.”\textbackslash n\textbackslash nPricing is not expected to change until two years from now, but by next year, the plans are already underway.\textbackslash n\textbackslash nTo start with, Winport will be delivered in next-generation shipping containers to clients: “Everybody wants to buy consoles in a box,” Assoui said. “They want a larger console and a lot more access to the music library and other libraries.”\textbackslash n\textbackslash nInside the K-8 is the machine learning engine.\textbackslash n\textbackslash nThe Serverless technology, which produces raw Google Image and has an AI model in the shape of a movie, has the potential to turn those calculations into real-time data for mobile augmented-reality interfaces and services. But the big challenge is what the machine learning system can be capable of.\textbackslash n\textbackslash n“We have a massive range of examples,” Assoui said. “It can be useful if we can do this in reality.”\textbackslash n\textbackslash nA K-8 can also be a platform for proof-of-concept, as it gives you the idea of a lab-like enterprise. That’s the kind of work that could already be done in a remotely controlled lab, thanks to robot factories, for example.\textbackslash n\textbackslash nThere’s a chance this will be Google’s next big product, if that is the case, said Don Schaefer, founder of the Evoliant developer group in Berkeley.\textbackslash n\textbackslash n“Basically building something gives it an advantage rather than just building a case that says, ‘Hey, build this,’” Schaefer said.\textbackslash n\textbackslash nThose advantages could not only be applicable to smartphone products, he said, but to big,}

\end{document}